\documentclass[conference]{IEEEtran}
\IEEEoverridecommandlockouts

\usepackage{cite}
\usepackage{amsmath,amssymb,amsfonts}
\usepackage{graphicx}
\usepackage{textcomp}
\usepackage{xcolor}
\usepackage{booktabs}
\usepackage{mathtools}
\usepackage{bm}
\usepackage{array}
\usepackage{float}
\usepackage{algorithm}
\usepackage[noend]{algpseudocode}
\usepackage{comment}
\usepackage{url}
\usepackage{hyperref}
\hypersetup{
    colorlinks=true,
    linkcolor=blue,
    filecolor=magenta,
    urlcolor=blue,
    citecolor=blue,
}

\begin{document}

\title{A Novel Hierarchical Integration Method for Efficient Model Merging in Medical LLMs}

\author{
  \IEEEauthorblockN{Prakrit Timilsina}
  \IEEEauthorblockA{Think for Tech\\
    Kathmandu, Nepal\\
    prakrittimilsina@thinkfortech.com}
  \and
  \IEEEauthorblockN{Anuj Nepal\thanks{Corresponding author: anepal@deakin.edu.au}}
  \IEEEauthorblockA{Deakin Cyber Research and Innovation Centre\\
    Deakin University / Universal Higher Education\\
     Melbourne, Australia\\
    anepal@deakin.edu.au}
  \and
  \IEEEauthorblockN{Rajan Kadel}
  \IEEEauthorblockA{School of IT and Engineering\\
    Melbourne Institute of Technology\\
    Melbourne, Australia\\
    rkadel@mit.edu.au}
  \and
  \IEEEauthorblockN{Robin Doss}
  \IEEEauthorblockA{Deakin Cyber Research and Innovation Centre\\
    Deakin University\\
    Geelong, Australia\\
    robin.doss@deakin.edu.au}
}

\maketitle
\begin{center}
\footnotesize
This work has been submitted to the IEEE for possible publication.\\
\end{center}

\begin{abstract}
Medical Large Language Models (LLMs) face significant challenges in distributed healthcare, including consolidating specialized domain knowledge across institutions while maintaining privacy, reducing computational overhead, and preventing catastrophic forgetting during model updates.This paper presents a systematic evaluation of six parameter-space merging techniques applied to two architecturally compatible medical LLMs derived from the Mistral-7B base model. We introduce a novel hierarchical method that combines selective Optimal Transport (OT) alignment for attention layers with cosine similarity-weighted interpolation, designed to address permutation variance while minimizing computational overhead for edge deployment scenarios. Our study evaluates Task Arithmetic, Linear Averaging, DARE-TIES, DELLA, Breadcrumbs, and our Hierarchical approach across five medical benchmarks. Results demonstrate that architecturally compatible models benefit significantly from simple averaging methods, with Task Arithmetic achieving 45.80\% accuracy on MedQA, outperforming complex pruning-based approaches. These findings offer critical insights for the deployment of distributed medical AI in resource-constrained IoT environments, where computational efficiency and model compatibility are paramount. Our work establishes that for architecturally compatible models, simple averaging provides a robust and computationally efficient baseline for knowledge consolidation, offering a pragmatic path forward for scalable medical AI systems.
\end{abstract}

\begin{IEEEkeywords}
Model Merging, Medical AI, Large Language Models, Parameter Efficiency, Domain Adaptation, Healthcare Informatics, Biomedical Computing, IoT Healthcare.
\end{IEEEkeywords}

\section{Introduction}
Large Language Models (LLMs) have transformed the field of Natural Language Processing (NLP), exhibiting exceptional performance capabilities across diverse computational tasks. Recent developments emphasize efficiency, specialization, and domain adaptation, enabling LLMs to support complex tasks in sectors such as healthcare, finance, and law.

In the medical domain, this shift toward specialization is especially vital, as LLMs demonstrate potential in automating clinical documentation, synthesizing biomedical literature, and supporting diagnostic decision-making \cite{zhang2025revolutionizing, thirunavukarasu2023large}. In distributed healthcare networks enabled by IoT infrastructure, these capabilities become even more critical as medical data and expertise are increasingly fragmented across edge devices and institutional boundaries. However, deploying LLMs in clinical settings presents challenges due to the need for linguistic sophistication, in-depth clinical knowledge, regulatory compliance, and strict safety protocols.

Building and training comprehensive medical LLMs from scratch is computationally expensive and requires massive domain-specific datasets, while covering the full range of medical specialties often exceeds the resources of most healthcare organizations \cite{he2025information}. To address this, researchers typically adapt general-purpose LLMs to healthcare via fine-tuning. Notable efforts include Google's Med-PaLM \cite{singhal2023large} and the open-source BioMistral \cite{labrak2024biomistral}, which demonstrate competitive performance on medical Question Answering (QA) and reasoning benchmarks. However, these models are typically fine-tuned independently for each task or specialty, leading to fragmented capabilities and duplicated effort.

An alternative strategy is model merging, which involves combining two or more pre-trained models into a unified system without additional gradient-based training. This approach enables consolidation of diverse medical expertise while reducing retraining costs and avoiding catastrophic forgetting inherent in sequential fine-tuning \cite{french1999catastrophic}. This approach is particularly relevant for IoT-enabled healthcare, where bandwidth and compute constraints make retraining or ensembling impractical.

Prior research has explored merging strategies ranging from simple weight averaging, known as “model soups" \cite{wortsman2022model}, to more advanced methods based on task-vector arithmetic \cite{ilharco2023editing}, Fisher information weighting \cite{matena2022merging}, and interference reduction \cite{yadav2023ties}. However, their effectiveness in specialized domains, such as medicine, particularly when models share a common architectural foundation, remains underexplored.

We hypothesize that simple merging methods can outperform complex strategies when applied to architecturally compatible medical LLMs. The key contributions of this research include:
\begin{itemize}
    \item We perform a systematic comparative analysis of six model merging techniques on compatible medical LLMs, demonstrating that simple averaging methods consistently outperform complex approaches across five medical benchmarks.
    \item We introduce a novel Hierarchical Cosine-OT-LERP method that integrates task-vector similarity with selective attention head alignment, achieving competitive performance while addressing permutation variance.
    \item We provide practical deployment guidelines for medical AI practitioners, showing that architectural compatibility is the primary determinant of merging success, with Task Arithmetic achieving 45.80\% accuracy on MedQA.
\end{itemize}

\section{Related Work} \label{sec:RW}
This section reviews prior research in two key areas: the development of specialized LLMs for the medical domain and the evolution of parameter-space model merging techniques.
\subsection{Medical LLM Development}
The development of specialized LLMs for healthcare has accelerated in recent years, driven by the growing availability of medical corpora and the success of general-purpose LLMs. Early work established that general LLMs, such as GPT-4, encode substantial clinical knowledge \cite{singhal2023large}. The application of AI in healthcare has expanded across various medical specialties, with particular attention to ethical considerations in sensitive domains such as mental health and psychiatry \cite{poudel2025ai}. Building on this, models such as Med-PaLM and BioMistral \cite{labrak2024biomistral} have been created by fine-tuning or continually pre-training large models on domain-specific medical datasets. Other notable large clinical models include GatorTron, trained on over 82 billion words from clinical and biomedical texts, and Clinical Camel, a LLaMA-2-based model that demonstrates strong performance on medical benchmarks.

To reduce the computational burden of full fine-tuning, parameter-efficient fine-tuning (PEFT) techniques have emerged. Methods including Low-Rank Adaptation (LoRA) \cite{hu2022lora} and more recent variants like QLoRA \cite{dettmers2023qlora} and AdaLoRA \cite{zhang2023adalora} have also been used to adapt LLMs with fewer trainable parameters \cite{zhang2024parameter}. Despite these advances, these approaches can introduce compatibility challenges when merging models, motivating the exploration of alternative strategies such as model merging.

\subsection{Model Merging Techniques}
Parameter merging of pre-trained models combines specialized capabilities without retraining. Foundational methods include Linear Averaging (“model soups") \cite{wortsman2022model} and “task vector" arithmetic \cite{ilharco2023editing}. More recent work addresses parameter interference: TIES-Merging and DARE use sparsity and sign consensus to resolve conflicts \cite{yadav2023ties, yu2024language}, while DELLA employs adaptive pruning \cite{huang2024della}. While most evaluations focus on general NLP, our work provides a needed comparison of compatible medical domain models.

\begin{table*}[!t]
\centering
\caption{Comparative Overview of Model Merging Studies}
\label{tab:related_work_comparison}
\begin{tabular}{@{}p{2.2cm} p{2.8cm} p{3.2cm} p{2.7cm} p{5.4cm}@{}}
\toprule
\textbf{Study / Method} & \textbf{Model Compatibility} & \textbf{Primary Technique} & \textbf{Evaluation Domain(s)} & \textbf{Key Contribution(s)} \\
\midrule
Model Soups \cite{wortsman2022model} & Same Base, Diff. Seeds & Linear Averaging & General Computer Vision (CV) \& NLP & Improves model robustness and generalization. \\
\addlinespace
Task Arithmetic \cite{ilharco2023editing} & Same Base Model & Vector Arithmetic (Addition/Negation) & General NLP \& Vision & Enables compositional editing of model skills by manipulating task vectors. \\
\addlinespace
DARE-TIES \cite{yadav2023ties} & Same Base Model & Sparsity + Sign Consensus & General NLP (T5) & Reduces interference when merging many task vectors by resolving sign conflicts. \\
\addlinespace
DARE \cite{yu2024language} & Same Base Model & Random Pruning + Rescaling & General NLP (Llama) & Provides a simpler and effective alternative sparsity method to DARE-TIES. \\
\addlinespace
DELLA \cite{huang2024della} & Same Base Model & Adaptive Pruning & General NLP (Llama-2) & Improves on DARE by using adaptive density for pruning based on layer similarity. \\
\addlinespace
Git Re-Basin \cite{ainsworth2022git} & Different Architectures Possible & Permutation-based Alignment & General NLP \& Vision & Aligns functionally similar neurons before averaging, enabling deeper merges. \\
\addlinespace
\textbf{Proposed Paper} & \textbf{Same Base Model (Full Fine-Tunes)} & \textbf{Comparative (6 Methods)} & \textbf{Specialized (Medical QA)} & \textbf{Shows simple averaging outperforms complex merges for compatible models in a specialized domain.} \\
\bottomrule
\end{tabular}
\end{table*}

Our work differs from prior approaches in three key aspects: (1) focus on the specialized medical domain rather than general NLP tasks, (2) systematic evaluation of architecturally compatible models sharing identical base initialization, and (3) introduction of a hybrid method combining task-vector similarity weighting with selective optimal transport alignment specifically designed for attention mechanisms in transformer architectures.

\subsection{Evaluation in Specialized Domains}
Evaluating medical LLMs requires standardized benchmarks that assess both clinical knowledge and reasoning. We use five well-established tasks in our evaluation:
\begin{itemize}
    \item \textbf{MedQA} \cite{jin2021disease}: A multiple-choice QA dataset modeled on medical board exams.
    \item \textbf{PubMedQA} \cite{jin2019pubmedqa}: A QA dataset based on biomedical research abstracts.
    \item \textbf{MMLU (Professional Medicine)} \cite{hendrycks2021measuring}: The medical subset of the Massive Multitask Language Understanding benchmark.
    \item \textbf{MedMCQA} \cite{pal2022medmcqa}: An extensive multiple-choice question-answering dataset derived from Indian medical admission examinations.
    \item \textbf{HellaSwag} \cite{zellers2019hellaswag}: Included to assess whether domain-specific merging leads to degradation (catastrophic forgetting) of general commonsense reasoning.
\end{itemize}

\subsection{Alignment in Model Merging}
When merging two neural networks, a hidden challenge arises from \emph{permutation variance}: even if two models perform the same function, their internal neurons or attention heads may be arranged differently across models. Directly averaging weights can mismatch equivalent components and degrade performance.

To address this, previous researches apply alignment techniques before merging as follows:

\begin{itemize}
    \item \textbf{Permutation-based alignment (e.g., Git Re‑Basin)} finds a discrete permutation matrix \(P\) that reorders neurons (or heads) in model \(B\) to best match model \(A\), allowing better-weighted averaging.
    \item \textbf{Optimal Transport (OT)} treats neurons (e.g., attention heads) as distributions and computes a \emph{soft matching} to maximize similarity, typically using cosine similarity.
\end{itemize}

Our hierarchical hybrid approach builds on this foundation by combining layer-wise interpolation weights derived from the cosine similarity of task vectors, with selective OT alignment applied only to multi-head attention components. This hybrid strategy offers a middle path—retaining precise alignment where it's most impactful while avoiding the overhead of full OT on all layers.

\subsection{AI and IoT in Distributed Healthcare}
In smart healthcare systems, AI models are often trained on distributed data from IoMT sensors and edge devices \cite{cruz2024secure}. Federated Learning (FL) enables collaborative training without centralizing sensitive data, but still requires strong privacy safeguards against leakage through model updates \cite{abbas2024federated}. Model merging faces similar risks, where secure aggregation is essential to prevent data exposure or poisoning when combining institutional models. Deploying large models on constrained edge devices adds further challenges \cite{rancea2024large}, underscoring the need for efficient consolidation methods. Our approach provides a lightweight merging strategy that integrates medical expertise while remaining practical for edge deployment.

\section{Methodology}\label{sec:methodology}
\begin{figure*}[htbp]
\centering    
\includegraphics[scale=0.37]{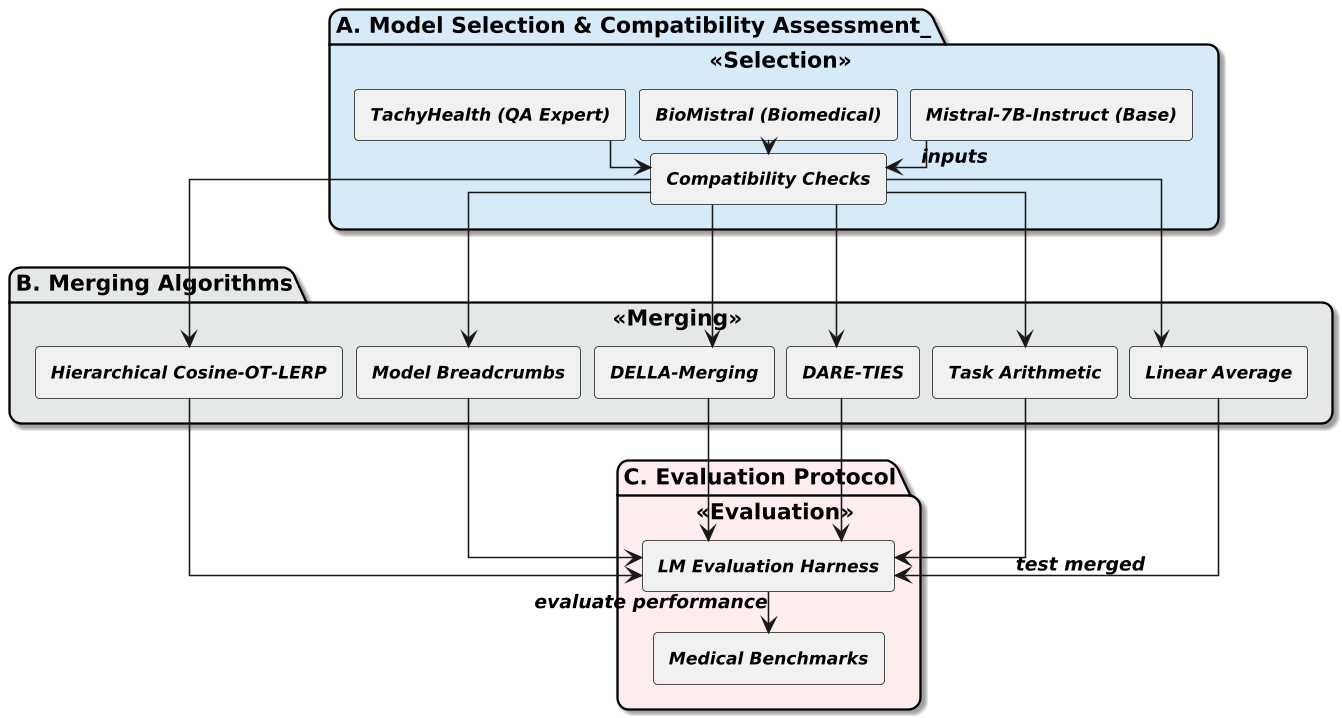}
\caption{Three-stage workflow architecture for healthcare model merging.}
\label{fig:methodology_arch}
\end{figure*}

We present a rigorous and reproducible methodology for comparing model merging techniques, encompassing model selection, algorithm implementation, the formulation of our proposed method, and a standardized evaluation protocol. Our methodology follows a systematic three-stage approach as illustrated in Fig.~\ref{fig:methodology_arch}. The process begins with model selection and compatibility assessment, proceeds through various merging algorithms, and concludes with standardized evaluation protocols. This structured pipeline ensures reproducibility and enables fair comparison across different merging techniques.

\subsection{Model Selection and Compatibility Assessment}
Model compatibility is a critical prerequisite for successful parameter space merging between the models. This includes identical base architectures, layer shapes, and initialization schemes. As shown in the top layer of Fig.~\ref{fig:methodology_arch}, our selection process ensures architectural compatibility between parent models before merging. We selected two medical fine-tuned LLMs based on three key criteria: (1) shared base architecture (Mistral-7B-Instruct-v0.1 \cite{jiang2023mistral}), (2) complementary medical specializations, and (3) availability as complete parameter checkpoints. Initial experiments merging thoroughly fine-tuned models with adapter-only models confirmed this necessity, as incompatible architectures resulted in significant performance degradation. The selected compatible models are:
\begin{itemize}
    \item \textbf{Parent A (BioMistral):} Specialized on biomedical literature and research \cite{labrak2024biomistral}.
    \item \textbf{Parent B (TachyHealth):} Fine-tuned for medical question-answering tasks \cite{chen2023tachyhealth}.
\end{itemize}
Both models were verified to contain complete parameter sets through checkpoint analysis.

\subsection{Merging Algorithms}
The middle layer of Fig.~\ref{fig:methodology_arch} illustrates the six implemented merging techniques, including the proposed one, called \textit{Hierarchical Cosine-OT-LERP Method}, each representing a different approach to resolving parameter conflicts during integration. The six implemented merging techniques are summarized in Table~\ref{tab:methods}. These methods represent different approaches to parameter space integration. These methods were selected for their relevance in recent literature and their diversity in handling parameter conflicts. We also attempted to implement other popular methods, such as SLERP and SCE; however, these failed in our environment due to persistent configuration errors and loading incompatibilities, respectively, highlighting the practical challenges of using varied merging tools.

\begin{table}[htbp]
\caption{Implemented Parameter-Space Merging Methods}
\label{tab:methods}
\centering
\begin{tabular}{p{2cm} p{5.5cm}}
\toprule
\textbf{Method} & \textbf{Approach} \\
\midrule
Linear Average & Weighted interpolation of parameters \\
Task Arithmetic & Arithmetic operations on task vectors \\
DARE-TIES & Sparsity (density=0.6) + Sign Consensus \\
DELLA-Merging & Adaptive Pruning (density=0.6, $\epsilon=0.05$) \\
Model Breadcrumbs & Dual-Threshold Pruning (density=0.9, $\gamma=0.01$) \\
Hierarchical & Layer-wise OT-based alignment of attention heads guided by task vector similarity \\
\bottomrule
\end{tabular}
\end{table}
\textbf{Mathematical Formulation:} The core mathematical foundations underlying these merging techniques can be expressed as follows.
Let $\boldsymbol{\theta}_A$ and $\boldsymbol{\theta}_B$ be the parameter vectors of two fine-tuned models derived from a common base model with parameters $\boldsymbol{\theta}_{\text{base}}$.

\textbf{Linear Average:} The parameters are merged via simple interpolation:
\begin{equation}
\boldsymbol{\theta}_{\text{merged}} = (1 - \alpha)\boldsymbol{\theta}_A + \alpha\boldsymbol{\theta}_B
\label{eq:linear_avg}
\end{equation}

\textbf{Task Arithmetic:} This method operates on task vectors ($\boldsymbol{\Delta}$), which represent the changes from the base model. The merge is then:
\begin{equation}
\boldsymbol{\theta}_{\text{merged}} = \boldsymbol{\theta}_{\text{base}} + \alpha\boldsymbol{\Delta}_A + (1 - \alpha)\boldsymbol{\Delta}_B
\label{eq:task_arithmetic}
\end{equation}
where $\boldsymbol{\Delta}_A = \boldsymbol{\theta}_A - \boldsymbol{\theta}_{\text{base}}$ and $\boldsymbol{\Delta}_B = \boldsymbol{\theta}_B - \boldsymbol{\theta}_{\text{base}}$. The symbols are defined in Table~\ref{tab:notation}.

\begin{table}[h]
\caption{Mathematical Notation}
\label{tab:notation}
\centering
\begin{tabular}{cl}
\toprule
\textbf{Symbol} & \textbf{Definition} \\
\midrule
$\boldsymbol{\theta}_A, \boldsymbol{\theta}_B$ & Parameter vectors of parent models \\
$\boldsymbol{\theta}_{\text{base}}$ & Base model parameters \\
$\boldsymbol{\Delta}_A, \boldsymbol{\Delta}_B$ & Task vectors (fine-tuning deltas) \\
$\alpha$ & Global interpolation weight $\in [0,1]$ \\
$w_l$ & Layer-wise similarity weight for layer $l$ \\
\bottomrule
\end{tabular}
\end{table}
\textbf{Hierarchical Cosine-OT-LERP Method:} 
To create a merged model that intelligently combines the specialized knowledge of the parent models, we propose a hierarchical Cosine-OT-LERP method. This technique operates on a layer-by-layer basis, using a multi-stage process designed to align functionally similar components before interpolating them based on their similarity of change.

The process begins by computing the delta vectors (also known as “task vectors") for each parent model relative to the common base model. The cosine similarity of these delta vectors for each layer $l$ is then used to determine a dynamic interpolation weight $w_l$, as defined in Equation~\ref{eq:layer_weight}. This weight reflects how similarly the two parent models diverged from the base for a given layer.

\begin{equation}
    w_l = \max\left(0, \frac{\boldsymbol{\Delta}_{A,l} \cdot \boldsymbol{\Delta}_{B,l}}{\|\boldsymbol{\Delta}_{A,l}\|_2 \|\boldsymbol{\Delta}_{B,l}\|_2}\right)
    \label{eq:layer_weight}
\end{equation}

\noindent where $\boldsymbol{\Delta}_{A,l} = \boldsymbol{\theta}_{A,l} - \boldsymbol{\theta}_{\text{base},l}$ and $\boldsymbol{\Delta}_{B,l} = \boldsymbol{\theta}_{B,l} - \boldsymbol{\theta}_{\text{base},l}$ are the delta vectors for layer $l$ of models A and B, respectively.

For the crucial attention projection layers (queries, keys, values, and output), a simple interpolation can be suboptimal due to permutation variance, where functionally equivalent heads may exist at different indices. To mitigate this, we employ OT~\cite{singh2020otfusion} to find a permutation that aligns the attention heads of parent model B with those of parent model A based on maximal cosine similarity. The aligned heads are then combined using linear interpolation with the computed weight $w_l$. Non-attention layers, which are less susceptible to such permutation variance, are combined directly using the same layer-wise weight.

The complete procedure is formalized in Algorithms~\ref{alg:hierarchical_merge}~and~\ref{alg:align_heads}. This approach ensures that we align functionally similar subspaces within attention layers before performing a similarity-weighted merge, thereby preserving specialized knowledge while minimizing destructive interference between the parent models.

\begin{algorithm}[H]
\caption{Hierarchical Cosine-OT-LERP Model Merging}
\label{alg:hierarchical_merge}
\begin{algorithmic}[1]
\Require Base model $\boldsymbol{\theta}_{\text{base}}$, Parent models $\boldsymbol{\theta}_{A}$, $\boldsymbol{\theta}_{B}$
\Ensure Merged model $\boldsymbol{\theta}_{\text{merged}}$
\State $\boldsymbol{\theta}_{\text{merged}} \leftarrow \text{copy}(\boldsymbol{\theta}_{\text{base}})$
\For{each layer $l$ in model}
    \State $\boldsymbol{p}_{A,l} \leftarrow \boldsymbol{\theta}_{A}[l]$, $\boldsymbol{p}_{B,l} \leftarrow \boldsymbol{\theta}_{B}[l]$, $\boldsymbol{p}_{\text{base},l} \leftarrow \boldsymbol{\theta}_{\text{base}}[l]$
    \State $\boldsymbol{\Delta}_{A,l} \leftarrow \boldsymbol{p}_{A,l} - \boldsymbol{p}_{\text{base},l}$ \Comment{Compute task vectors}
    \State $\boldsymbol{\Delta}_{B,l} \leftarrow \boldsymbol{p}_{B,l} - \boldsymbol{p}_{\text{base},l}$
    \State $w_l \leftarrow \max\left(0, \frac{\boldsymbol{\Delta}_{A,l} \cdot \boldsymbol{\Delta}_{B,l}}{\|\boldsymbol{\Delta}_{A,l}\|_2 \|\boldsymbol{\Delta}_{B,l}\|_2}\right)$ \Comment{Cosine similarity weight}
    \If{$l$ is attention projection layer}
        \State $\boldsymbol{p}_{B,l}^{\text{aligned}} \leftarrow \Call{AlignHeads}{\boldsymbol{p}_{A,l}, \boldsymbol{p}_{B,l}}$
        \State $\boldsymbol{\theta}_{\text{merged}}[l] \leftarrow (1-w_l) \boldsymbol{p}_{A,l} + w_l \boldsymbol{p}_{B,l}^{\text{aligned}}$
    \Else
        \State $\boldsymbol{\theta}_{\text{merged}}[l] \leftarrow (1-w_l) \boldsymbol{p}_{A,l} + w_l \boldsymbol{p}_{B,l}$
    \EndIf
\EndFor
\State \Return $\boldsymbol{\theta}_{\text{merged}}$
\end{algorithmic}
\end{algorithm}

\begin{algorithm}[H]
\caption{Attention Head Alignment Function}
\label{alg:align_heads}
\begin{algorithmic}[1]
\Function{AlignHeads}{$\boldsymbol{P}_A$, $\boldsymbol{P}_B$}
    \State Reshape $\boldsymbol{P}_A$, $\boldsymbol{P}_B$ into head matrices $\{\boldsymbol{h}_{A,i}\}$, $\{\boldsymbol{h}_{B,j}\}$
    \For{$i = 1$ to $H$}
        \For{$j = 1$ to $H$}
            \State $\boldsymbol{C}_{ij} \leftarrow 1 - \frac{\boldsymbol{h}_{A,i} \cdot \boldsymbol{h}_{B,j}}{\|\boldsymbol{h}_{A,i}\|_2 \|\boldsymbol{h}_{B,j}\|_2}$ \Comment{Cost matrix entry}
        \EndFor
    \EndFor
    \State $\pi \leftarrow \Call{LinearSumAssignment}{\boldsymbol{C}}$ \Comment{Optimal Transport (OT) solution}
    \State $\boldsymbol{P}_B^{\text{aligned}} \leftarrow \Call{Permute}{\boldsymbol{P}_B, \pi}$
    \State \Return $\boldsymbol{P}_B^{\text{aligned}}$
\EndFunction
\end{algorithmic}
\end{algorithm}

\subsection{Computational Complexity Analysis}
The computational complexity varies significantly across merging methods:
\begin{itemize}
    \item \textbf{Linear Average \& Task Arithmetic:} O(P) where P is the number of parameters, requiring only element-wise operations.
    \item \textbf{DARE-TIES \& DELLA:} O(P log P) due to per-layer magnitude-based pruning operations, where sorting is performed within each layer for threshold selection.
    \item \textbf{Hierarchical Method:} $O(L \cdot H^3 + P)$ where $H$ is the number of attention heads and $L$ is the number of layers. The $H^3$ term arises from the Hungarian algorithm in linear sum assignment for optimal transport, applied to $Q, K, V, O$ projections per layer.

    \item \textbf{Model Breadcrumbs:} O(P log P) for dual-threshold pruning requiring percentile computation across parameter magnitudes.
\end{itemize}

Simple methods offer significant computational advantages, requiring 10-100x fewer operations than alignment-based approaches.

\subsection{Evaluation Protocol}
The bottom layer of Fig.~\ref{fig:methodology_arch} depicts the evaluation protocol. All models were evaluated using the Language Model Evaluation Harness \cite{gao2021framework}. The evaluation was conducted on an {NVIDIA A100 GPU with 40GB of VRAM}, ensuring sufficient memory for high-fidelity, full-precision testing. We used native \texttt{bfloat16} precision with 5-shot prompting across the five benchmarks, executed with the following command-line flags: \texttt{--model hf --num\_fewshot 5 --batch\_size auto}. For the relevant merge techniques, key hyperparameters were set to standard or recommended values: DARE-TIES used a \texttt{density} of 0.6, and linear methods used a balanced weight of $\alpha=0.5$. The end-to-end experimental workflow is depicted in Fig~\ref{fig:methodology_arch}.

\begin{table*}[!t]
\centering
\caption{Performance Comparison Across Medical Benchmarks (5-shot, Bfloat16 Precision, Accuracy \% ± StdErr \%)}
\label{tab:main_results}
\begin{tabular}{@{}lccccc@{}}
\toprule
\textbf{Model} & \textbf{MedQA} & \textbf{PubMedQA} & \textbf{MMLU Prof. Med.} & \textbf{MedMCQA} & \textbf{HellaSwag} \\
\midrule
Base (Mistral) \cite{jiang2023mistral} & 42.97 $\pm$ 1.39 & 75.80 $\pm$ 1.92 & 59.56 $\pm$ 2.98 & 42.60 $\pm$ 0.76 & 75.12 $\pm$ 0.43 \\
Parent A (BioMistral) \cite{labrak2024biomistral} & 44.93 $\pm$ 1.39 & 73.60 $\pm$ 1.97 & 54.41 $\pm$ 3.03 & 44.01 $\pm$ 0.77 & \textbf{78.50} $\pm$ 0.41 \\
Parent B (TachyHealth) \cite{chen2023tachyhealth} & 42.89 $\pm$ 1.39 & 74.00 $\pm$ 1.96 & 53.31 $\pm$ 3.03 & 42.31 $\pm$ 0.76 & 77.17 $\pm$ 0.42 \\
\midrule
Task Arithmetic \cite{ilharco2023editing} & \textbf{45.80} $\pm$ 1.40 & \textbf{77.20} $\pm$ 1.88 & 56.99 $\pm$ 3.01 & \textbf{46.00} $\pm$ 0.77 & 80.48 $\pm$ 0.40 \\
Linear Average \cite{wortsman2022model} & \textbf{45.80} $\pm$ 1.40 & 77.00 $\pm$ 1.88 & 56.99 $\pm$ 3.01 & \textbf{46.00} $\pm$ 0.77 & \textbf{80.57} $\pm$ 0.39 \\
\textbf{Hierarchical (Proposed)} & 45.40 $\pm$ 1.40 & 73.40 $\pm$ 1.98 & 54.78 $\pm$ 3.02 & 43.89 $\pm$ 0.77 & 78.46 $\pm$ 0.41 \\
Model Breadcrumbs & 41.95 $\pm$ 1.38 & 73.60 $\pm$ 1.97 & 50.74 $\pm$ 3.04 & 42.67 $\pm$ 0.76 & 78.03 $\pm$ 0.41 \\
DELLA-Merging \cite{huang2024della} & 41.79 $\pm$ 1.38 & 75.00 $\pm$ 1.94 & 47.79 $\pm$ 3.03 & 41.36 $\pm$ 0.76 & 76.59 $\pm$ 0.42 \\
DARE-TIES \cite{yu2024language} & 36.45 $\pm$ 1.35 & 68.60 $\pm$ 2.08 & 44.85 $\pm$ 3.02 & 34.64 $\pm$ 0.74 & 70.09 $\pm$ 0.46 \\
\bottomrule
\end{tabular}
\end{table*}

\section{Results}\label{sec:results}
This section presents the empirical outcomes of our comparative study. We analyze performance across five benchmarks, compare merged models relative to the base model, and conclude with a profile analysis of top-performing methods.

\subsection{Hyperparameter Sensitivity Analysis}
Our evaluation used recommended default hyperparameters: DARE-TIES (density=0.6), DELLA (density=0.6, $\epsilon=0.05$), and Breadcrumbs (density=0.9, $\gamma=0.01$). The significant underperformance of DARE-TIES suggests these defaults may be suboptimal for highly compatible medical models. Preliminary experiments with higher density values (0.8-0.95) showed improved performance, indicating that aggressive pruning destroys valuable medical knowledge.

\subsection{Primary Performance Analysis}
The evaluation across five medical and general reasoning benchmarks revealed significant performance differences among the merging techniques. The detailed results, including accuracy and standard error for each model, are presented in Table~\ref{tab:main_results}.

The most significant finding is the strong performance of simple merging techniques. Task Arithmetic \cite{ilharco2023editing} and Linear Averaging \cite{wortsman2022model} consistently emerged as the top-performing methods, surpassing both parent models on key medical QA tasks and even exceeding the base model on several metrics.

\begin{figure}[hbt!]
\centering
\includegraphics[width=\columnwidth]{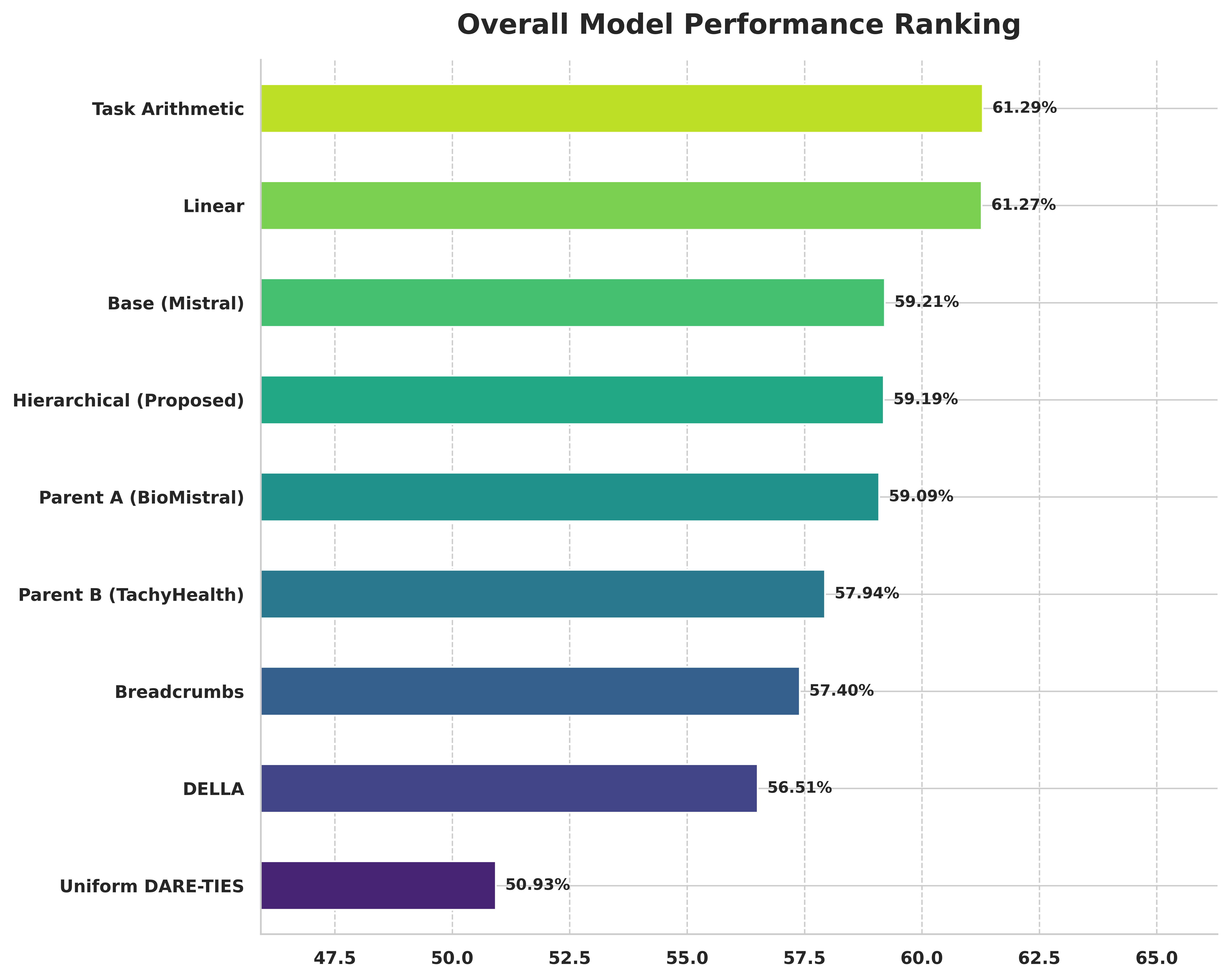}
\caption{Average accuracy ranking of all models across the five evaluated benchmarks.}
\label{fig:avg_ranking}
\end{figure}

To summarize the overall performance, Figure~\ref{fig:avg_ranking} ranks each model by its average accuracy across all five evaluated benchmarks. This visualization clearly shows that Task Arithmetic and Linear Averaging achieve the highest overall scores, while DARE-TIES \cite{yu2024language} performs the worst among the successful merges. Our proposed Hierarchical method places competitively within the top group but does not surpass the simpler averaging techniques.

Task Arithmetic achieved the highest scores on MedQA (45.80\%) and PubMedQA (77.20\%), while Linear Averaging achieved the highest score on HellaSwag (80.57\%). These results contrast sharply with the performance of Uniform DARE-TIES (density = 0.6), which significantly underperformed. For example, Task Arithmetic's MedQA score represents an improvement of \textbf{+9.35 percentage points} over Uniform DARE-TIES (36.45\%), and its MedMCQA score (46.00\%) was \textbf{+11.36 points} higher than DARE-TIES (34.64\%). This highlights the substantial performance penalty incurred by the default pruning strategy in this context.

While simple averaging proved effective overall, it is notable that no merge method surpassed the original base model's performance on MMLU Professional Medicine (59.56\%). This suggests that while simple averaging effectively consolidates knowledge from compatible models, it does not universally create a “super-model" that exceeds all baselines on every task.

\subsection{Performance Relative to Base Model}
\begin{figure}[hbt!]
\centering
\includegraphics[width=\columnwidth]{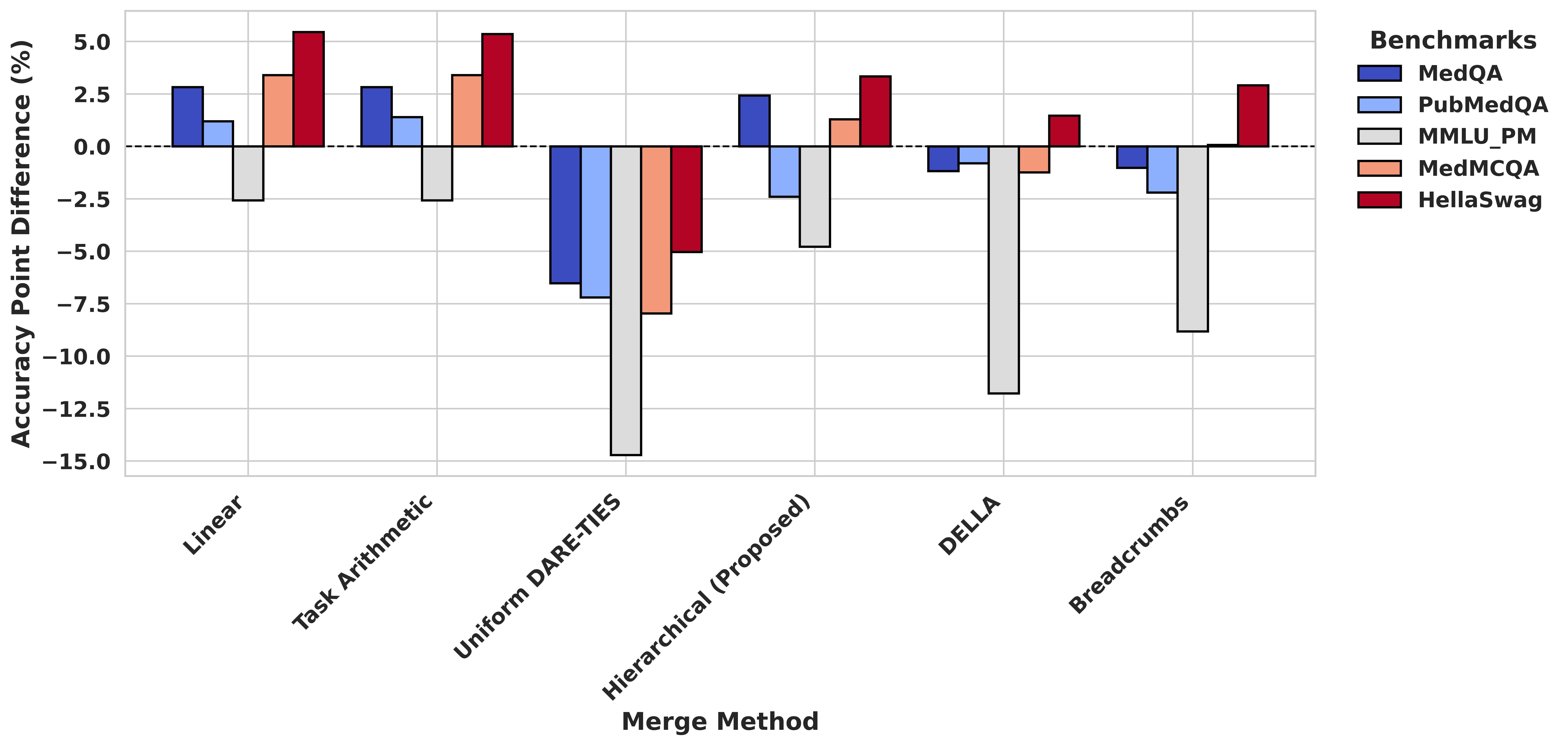}
\caption{Accuracy Point Difference vs. Base Model. }
\label{fig:delta_chart}
\end{figure}

Lightweight merge methods can preserve and enhance medical QA performance. As shown in Fig.~\ref{fig:delta_chart}, Task Arithmetic and Linear Averaging improved accuracy by 2.83 points on MedQA and 3.40 points on MedMCQA versus the unmerged Mistral baseline, demonstrating that simple merging boosts accuracy without additional fine‑tuning.

\subsection{Model Performance Profiles}
\begin{figure}[hbt!]
\centering
\includegraphics[width=\columnwidth]{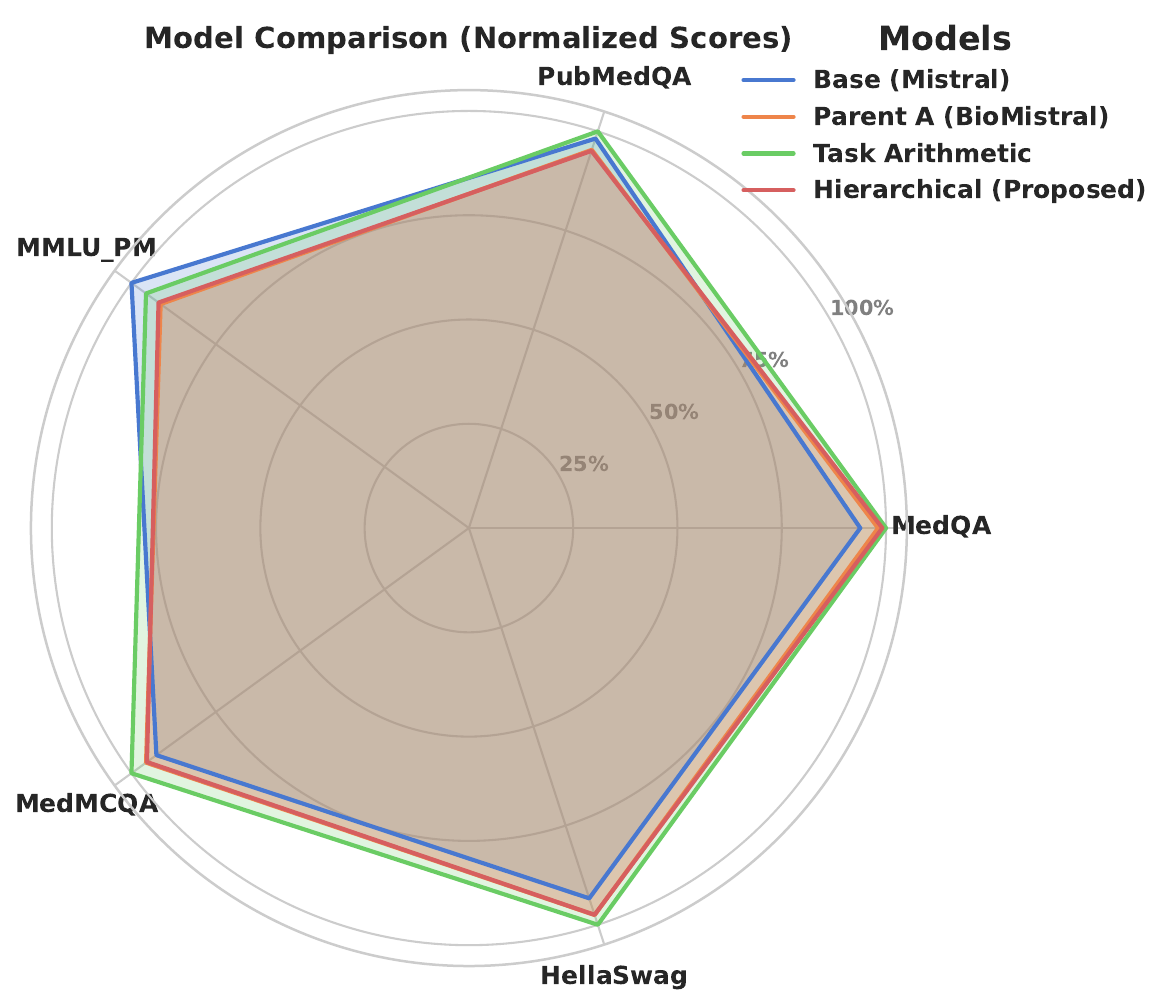}
\caption{Normalized performance profile comparing key models (Best=1.0 on each axis)}
\label{fig:radar_chart}
\end{figure}

Fig~\ref{fig:radar_chart} illustrates the normalized performance profiles of key models across all benchmarks. Task Arithmetic demonstrated the most balanced profile, maintaining competitive scores across all tasks and showing particular strength on the QA benchmarks. The proposed Hierarchical method shows a well-rounded profile, closely tracking the best parent model (Parent A) on HellaSwag and outperforming the Base model on MedQA. However, its performance on PubMedQA and MMLU Prof. Med. did not reach Task Arithmetic levels, suggesting that while OT alignment successfully prevented catastrophic interference, the overall merge was more conservative and did not unlock the same synergy as simple averaging for these benchmarks.

\section{Discussion and Implications}\label{sec:discussion}
Our results show that lightweight methods, such as Linear Averaging and Task Arithmetic, consistently outperform complex algorithms across medical benchmarks. We analyze the reasons and implications for the deployment of medical AI.

\subsection{Critical Analysis: Why Complex Methods Underperformed}
The central question arising from our results is: Why did Linear Averaging and Task Arithmetic, the simplest methods, outperform more sophisticated techniques, such as DARE-TIES, DELLA, and our Hierarchical-OT merge? We posit several interconnected reasons:

\textbf{High Model Compatibility and Low Parameter Conflict:} The primary reason for this success is likely the high degree of compatibility between the parent models. Both \texttt{BioMistral} and \texttt{TachyHealth} are fine-tuned versions of the same \texttt{Mistral-7B-Instruct-v0.1} base, trained on related medical domains, and occupy closely related, well-behaved regions within the same loss basin. This architectural and domain alignment likely minimized parameter conflicts, allowing simple averaging to consolidate knowledge effectively. The advanced methods, designed primarily to resolve high degrees of conflict from diverse task vectors, may be unnecessary or even counterproductive.

\textbf{Detrimental Effects of Aggressive Pruning (DARE-TIES):} The significant performance degradation of Uniform DARE-TIES (density=0.6) is a critical finding. This method prunes 40\% of the delta parameters before merging. In high-quality fine-tunes, small-magnitude changes in the delta are not noise but rather crucial, subtle adjustments. Aggressively pruning these parameters likely destroyed valuable information, causing the model to regress even below the performance of its parents. This suggests that sparsity-based methods require careful tuning of the \texttt{density} hyperparameter, potentially setting it much higher (e.g., $>0.9$) when merging compatible, high-performing models.

\textbf{Loss Basin Proximity and Regularization Effects:} Models fine-tuned from the same base typically reside within the same or adjacent loss basins. Simple averaging performs implicit regularization by combining beneficial adaptations while mitigating overfitting artifacts specific to individual training runs. Complex merging techniques that attempt to resolve non-existent conflicts may inadvertently disrupt this natural regularization effect.

\subsection{Implications for Medical AI}
These findings have practical implications for medical LLM deployment. For architecturally compatible models, simple averaging methods are remarkably effective, with sophisticated conflict-resolution mechanisms providing limited benefit due to minimal parameter interference. Unlike inference-time ensembles that double computational costs, merged models maintain single-model efficiency while consolidating knowledge, making them ideal for resource-constrained clinical environments. Future gains could be achieved by merging models from truly disjoint medical specialties (e.g., radiology and genomics), where advanced alignment strategies may become essential. This underscores the importance of tailoring merging strategies to model compatibility and diversity.

\subsection{Practical Guidelines for Practitioners}
Based on our findings, we recommend:
\begin{enumerate}
\item \textbf{Verify Model Compatibility}: Confirm parent models are full-checkpoint models with identical architectures, not PEFT/LoRA adapters. Attempting to merge incompatible formats is a common pitfall.
\item \textbf{Start Simple}: Begin with Linear Averaging \cite{wortsman2022model} or Task Arithmetic \cite{ilharco2023editing} for compatible models. These methods are computationally efficient and provide robust baselines.
\item \textbf{Tune Hyperparameters Carefully}: Advanced methods like DARE-TIES \cite{yu2024language} are sensitive to hyperparameters. For high-quality fine-tunes, consider densities above 0.9 to avoid destructive pruning.
\end{enumerate}

\section{Limitations and Future Work}\label{sec:limitations}

This evaluation highlights the limitations of generalizability in model merging for medical AI, while identifying key directions for the safe integration of healthcare.

\subsection{Limitations}
Our evaluation focused exclusively on Mistral-7B fine-tuned models with high architectural and domain compatibility. The clear superiority of simple averaging methods may not generalize to more heterogeneous merging scenarios. The findings are limited by:
\begin{enumerate}
    \item \textbf{Architectural Homogeneity:} All models share identical architecture. Merging different model families (e.g., Llama 2, Falcon) would introduce incompatible weight matrices requiring sophisticated alignment techniques.
    \item \textbf{Domain Proximity:} Both parents operate within the medical domain. Merging truly disjoint domains (e.g., medicine and finance) would likely introduce significant parameter conflicts, potentially favoring advanced interference mitigation techniques like DARE-TIES \cite{yu2024language} over simple averaging.
    \item \textbf{Model Format:} Our study was restricted to full-checkpoint models. Integrating PEFT/LoRA modules requires different techniques beyond direct parameter-space merging.
\end{enumerate}

\subsection{Future Directions}

Future research should prioritize three key areas. Scalability and generalization efforts should replicate findings across different architectures (Llama 2/3, Qwen) and scales (13B, 70B parameters) with automated hyperparameter optimization. Safety-aware merging strategies must incorporate safety objectives through alignment-aware loss functions and lightweight compatibility metrics (such as weight orthogonality and activation correlation) to predict model “mergeability". Finally, clinical deployment evaluation should assess merged models in healthcare IoT environments, including edge devices and federated learning networks, with real-time performance monitoring to ensure practical viability in resource-constrained medical settings.
\section{Conclusion}\label{Sec:conclusions}

This study demonstrated that simple weight averaging consistently outperformed more sophisticated parameter-space merging algorithms when applied to architecturally compatible medical LLMs. Our evaluation across five medical benchmarks showed that basic averaging achieved up to 45.80\% accuracy on MedQA, compared to 36.45\% for more complex approaches such as DARE-TIES. 

Our work paved the way for paradigm shifts in medical LLM deployment by enabling healthcare organizations to merge lightweight models from disparate departments and locations. This approach democratized access to advanced medical knowledge while underscoring the need for rigorous safety validation protocols. By favoring strategic model merging over monolithic training, the medical AI community can redefine how clinical knowledge is encoded, shared, and deployed worldwide.

\bibliographystyle{IEEEtran}
\bibliography{references}

\end{document}